# Leveraging the Interplay Between Syntactic and Acoustic Cues for Optimizing Korean TTS Pause Formation

Yejin Jeon[1], Yunsu Kim[3], Gary Geunbae Lee[1,2]

[1]Graduate School of Artificial Intelligence, POSTECH, Republic of Korea
[2]Department of Computer Science and Engineering, POSTECH, Republic of Korea
[3] aiXplain Inc. Los Gatos, CA, USA
{jeonyj0612, gblee}@postech.ac.kr, yunsu.kim@aixplain.com

**Abstract**

Contemporary neural speech synthesis models have indeed demonstrated remarkable proficiency in synthetic speech generation as they have attained a level of quality comparable to that of human-produced speech. Nevertheless, it is important to note that these achievements have predominantly been verified within the context of high-resource languages such as English. Furthermore, the Tacotron and FastSpeech variants show substantial pausing errors when applied to the Korean language, which affects speech perception and naturalness. In order to address the aforementioned issues, we propose a novel framework that incorporates comprehensive modeling of both syntactic and acoustic cues that are associated with pausing patterns. Remarkably, our framework possesses the capability to consistently generate natural speech even for considerably more extended and intricate out-of-domain (OOD) sentences, despite its training on short audio clips. Architectural design choices are validated through comparisons with baseline models and ablation studies using subjective and objective metrics, thus confirming model performance.

## 1. Introduction

Significant progress in the field of speech synthesis is largely credited to the emergence of seminal neural end-to-end (E2E) models such as Tacotron (Wang et al., 2017; Shen et al., 2018) and FastSpeech (Ren et al., 2019, 2021). Nevertheless, the predominant focus of research and empirical validations has been for high-resource languages like English. In contrast, the investigation and adaptation of contemporary neural text-to-speech (TTS) models to other languages pose substantial challenges, as it demands additional linguistic configurations (Yasuda et al., 2019; Xu et al., 2021) to accommodate the distinctive phonetic, morphological, and syntactic nuances inherent to each language.

In pursuit of achieving native-like fluency in a specific language with TTS, the identification of language-specific components assumes paramount significance. In the context of Korean, the strategic employment of pauses emerges as a critical element for effective conveyance of meaning, structure, and emphasis (Kang, 2010). As illustrated in Figure 1, the positioning of pauses within an utterance can dramatically metamorphose its intended message[1]. Hence, the precise prediction and placement of pauses within a sentence assume a critical role for Korean speech synthesis.

Pauses in speech can be broadly classified into two categories: punctuation-induced (Hieke et al., 1983) and respiratory pauses (Bailly and Gouvernayre, 2012). While the former aligns with clear speech segment boundaries that are marked by punctuation marks such as periods and commas, respiratory pauses prove to be more intricate; their positioning is substantially influenced by 1) linguistic organization encompassing grammar and syntax, as well as 2) acoustic cues like boundary tones, declination reset, and prepausal lengthening, which serve to connect adjacent speech segments. Towards the enhancement of verbal fluency in synthetic Korean speech, we concentrate on refining the modeling of respiratory pauses.

An intuitive branch of research attempts to improve pause predictions through explicit pause categorization and annotation. In Lu et al. (2019), all pause boundaries are manually annotated into either boundary or non-boundary tags. Similarly, Guo et al. (2019) categorizes pauses into four groups, and employ supplementary descriptors such as the lengths of surrounding intonational phrases to model inter-sentential and intra-sentential pausing. Yang et al. (2023) differentiates pauses according to duration with a Gaussian mixture model. However, these approaches are costly as they require manual annotation. Even if pause annotations are automated via classification, relying solely on duration as the primary deterministic criteria does not accurately mirror the complex variations observed in real-life speech.

| Verbal Stream | Transliteration |
|---|---|
| 아버지가 방에 들어가신다. | Dad is going into the <u>room</u>. |
| 아버지 가방에 들어가신다. | Going into <u>dad's bag</u>. **OR** Dad is going into the <u>bag</u>. |
| 오늘밤 나무 사 온다. | Will buy a <u>tree</u> tonight. |
| 오늘밤 나 무 사 온다. | I will buy a <u>radish</u> tonight. |

Figure 1: Cases where interpretations of synthetic utterances are influenced by different pausing locations. Corresponding segments are underlined.

---
[1]For both autoregressive and non-autoregressive TTS, we find that pauses do not necessarily align with inter-word spacing.

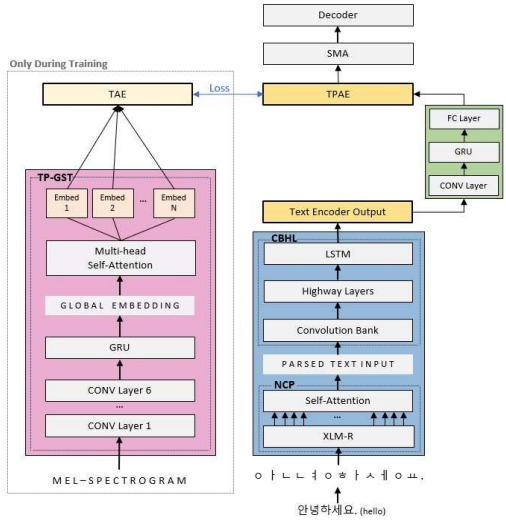

Figure 2: Architecture of the proposed model.

An alternative approach involves the incorporation of language-specific knowledge derived explicitly from textual inputs, specifically the inherent structural relationships between words. For instance, Bachenko et al. (1986) predicates their work on the premise that grammatical relationships dictate phrase boundary occurrences. Similarly, subsequent works adopt POS and a wordŠs relative hierarchical position in a sentence to approximate syntactic tree information (Guo et al., 2019). More recent studies calculate syntactic dependency distances between accented phrases (Kaiki et al., 2021), or utilize graph neural networks to extract syntax trees from text and encode linguistic relationships between word pairs (Liu et al., 2021). Yet, these studies only focus on the extraction of linguistic context solely from the textual modality. In fact, the formation of phrasal units concerning pauses is influenced by both syntactic (Grosjean et al., 1979; Jun, 1998) and acoustic attributes (Price et al., 1991; Beach, 1991). In light of this, we introduce a novel framework that harnesses both types of information in an unsupervised manner.

Another challenge pertains to the data-intensive nature of neural TTS systems, which necessitate extensive high-quality data for the synthesis of natural-sounding speech. This becomes particularly pronounced in the context of languages with limited data resources. For instance, while of high quality, the widely-used open-source Korean KSS (Park, 2018) dataset is considerably smaller than its English (Ito and Johnson, 2017), German (Müller and Kreutz, 2021), or Japanese (Kawai et al., 2004) counterparts. Addressing this constraint by curating larger datasets can swiftly escalate into a prohibitively expensive undertaking. Consequently, it becomes crucial to efficiently and accurately extract and learn prosodic features and utilize them to reliably synthesize natural speech, even for substantially longer sentences not encountered in open-source, but limited data.

Towards natural Korean speech generation, our contributions are as follows: 1) We apply a neural TTS system to Korean and focus on solving pausing errors, which greatly influence speech interpretation and quality. 2) We propose a novel framework to model syntactic and acoustic properties, enhancing Korean TTS performance. 3) We analyze which pausing patterns are encouraged when using syntactic or acoustic information, and motivate the interplay between different features. 4) Despite limited resources, our TaKOtron2-Pro model is able to accurately and robustly insert pauses within a given text, and generalize well for out-of-domain (OOD) sentences that are much complex and longer than those seen in the training data. This is done in an unsupervised manner without the need for supplementary annotations.

## 2. METHODOLOGY

In this section, we present TaKOtron2-Pro (Figure 2), which is a Korean autoregressive TTS system implemented on a baseline Tacotron2 model with stepwise monotonic attention[2] (He et al., 2019).

### 2.1. Input Units

The Korean writing system is an alphabetic syllabary; the basic pronunciation unit is a syllable, which is made up of two to four alphabetic characters. This means that the input to the TTS system can be processed either as a decomposition of individual letters or as a syllabic unit. We chose the former method for two reasons. First, Korean realizes an almost ideal 1:1 alignment between characters and phonemes (Taylor, 1980). Second, there are 11,172 possible syllabic representations in the Korean language, making it difficult to find or create a corpus that has all variants. Therefore, using characters is a much more viable option. An example of character-based text processing is illustrated in Figure 2. The Korean honorific version of *hello* is originally pronounced with 5 syllables, but is converted into a representation of 12 characters.

---
[2]We hereby refer to Tacotron2 with SMA as Tacotron2*

|  | MOS-S | MOS-L | WER-S | WER-L |
|---|---|---|---|---|
| Tacotron2* | 3.233 ± 0.07 | 2.900 ± 0.08 | 0.13821 | 0.23729 |
| FastSpeech2 | 3.267 ± 0.07 | 2.733 ± 0.07 | 0.13025 | 0.45024 |
| **TaKOtron2-Pro** | 3.467 ± 0.08 | 3.767 ± 0.07 | 0.12429 | 0.14689 |

Table 1: Comparison of the proposed model against baseline models. *S* and *L* refer to short and long sentence settings, respectively.

## 2.2. Syntactic Features

The significance of syntactic information can be attested by the task of oral reading; when reading out loud, readers conduct prosodic planning by looking a couple of characters ahead of the current word (Choi and Koh, 2009). This is to know where to pause next, creating a unit from the current reading position up to the next pause boundary.

To extract such local contextual information, we use a module similar to Lee et al. (2017) as part of the context encoder. This module is made up of 1-D **c**onvolution filter **b**anks, **h**ighway networks, and a bidirectional **L**STM, hereafter referred to as CBHL. Due to the various filter widths used in the convolution bank, CBHL is able to model unigrams, bigrams, trigrams, and up to N-grams of the input text sequence. Yet, this approach has a potential limitation; as readers only have a limited perceptual span, wrong predictions of the next constituent boundary can be made, resulting in incorrect prosody generation. In such cases, readers must go back and reproduce speech using the correct prosodic patterns. This suggests that global context (i.e., knowledge of all constituent boundary locations) is also important for natural speech generation.

To further incorporate this logic into our system, we use a pre-trained neural constituency parser (NCP) (Kitaev et al., 2019) made up of a pre-trained XLM-R (Conneau et al., 2020) and multi-head self-attention to parse a sentence into its respective constituents. In other words, when passing the initial TTS input character sequence to this parser, we identify all constituent boundaries so that the ending position of each non-terminal phrasal category such as the noun and verb phrase in an utterance is indicated by a special pipeline character. This newly processed text is then used as input to CBHL.

## 2.3. Acoustic Features

Listeners rely on acoustic cues in an utterance to discern meaning. This means that during speech generation, the cues that listeners use for speech perception such as major prosodic breaks should be taken into account for effective communication. Since this information cannot be learned from just contextual information, there have been many attempts to learn these prosodic features directly from audio. However, as they are difficult to annotate, unsupervised approaches like global style tokens (GST) (Wang et al., 2018) and variational autoencoders (Kingma and Welling, 2013) are commonly used to model these features as latent variables. Yet, a major drawback of these methods is that they require an auxiliary acoustic input during inference to maintain prosodic control. Selecting an audio with the preferred prosody then becomes an additional point for consideration.

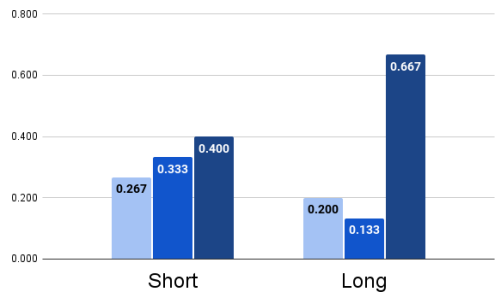

Figure 3: ABX results of short and long utterances synthesized by Tacotron2* (left), FastSpeech2 (middle), and TaKOtron2-Pro (right).

We predict the above pause-related acoustic features without requiring an auxiliary input during inference using TP-GST (Stanton et al., 2018). To do so, during training, acoustic input corresponding to input text is passed to a stack of six 2-D convolution layers, 128-unit GRU, and multi-head self-attention. Consequently, a representation for each acoustic input is generated, which we refer to as the target acoustic embedding (TAE) since it is designated as a target for the text encoder (Section 2.2) to predict. However, the output of the text encoder is a variable-length text embedding, so a fixed-length embedding is obtained by passing it through an additional CNN layer, a time-aggregating GRU, a fully connected layer, and a tanh activation. Although K fully connected layers can be used as in Stanton et al. (2018), we found no significant improvements from using multiple layers. As such, we simplify the model by using just one fully connected layer.

## 2.4. Training Loss

Our training approach employs multiple loss terms for training. Specifically, we utilize MSE mel-spectrogram reconstruction losses both before and after the Tacotron2* post-net and stop prediction stages. Additionally, we incorporate an L1 loss term between TAE and the final output of the text encoder (i.e., Text-Predicted Acoustic Embedding (TPAE)). To ensure effective training, we halt gradient flow to prevent text prediction errors from propagating backward through TP-GST layers. We empirically set the weight parameter ($\lambda$) for this mech-

| Type | Model | MOS | WER | ABX |
|---|---|---|---|---|
| S | CBHL | 3.400 ± 0.13 | 0.14204 | 0.133 |
| | CBHL + NCP | 3.367 ± 0.13 | 0.14102 | 0.200 |
| A | TP-GST | 3.600 ± 0.14 | 0.13909 | 0.267 |
| S + A | TaKOtron2-Pro | 3.750 ± 0.08 | 0.13559 | 0.400 |

Table 2: Ablations for syntactic and acoustic assimilation. *S* and *A* under the Type category refers to syntactic and acoustic features, respectively.

anism to 0.3, and the entire training objective is delineated as follows:

$$Loss_{Total} = Loss_{Mel} + Loss_{Gate} + \lambda \cdot Loss_{TP\text{-}GST} \quad (1)$$

## 3. Experimental Results

### 3.1. Training Setup

The training process of the entire TTS system involves separate training of the proposed TaKOtron2-Pro model and a MelGAN (Kumar et al., 2019) vocoder. The proposed model is trained using a batch size of 16 and one A100 GPU. Learning rate is annealed from 1e-3 to 5e-4, and then to 3e-4 every 50,000 iterations. All models are trained with the open source KSS dataset, which is made up of 22050 Hz audio files that are on average 2.38 seconds long. A 9:1 ratio is used to split the dataset into training and validation sets, respectively.

### 3.2. Experiments and Results

We compare our TaKOtron2-Pro model with representative baselines under two distinct conditions: 1) short sentence synthesis with audios matching the average length in the KSS dataset, and 2) synthesis of longer OOD sentences with an average of 22 words per sentence. 15 native Korean speakers rated the speech of various models with MOS and ABX metrics in which the former utilizes a 5 point Likert scale with 0.5 increments. Furthermore in all evaluations, we only included synthetic utterances that were devoid of any synthesis failure (e.g., incomprehensible noise) to ensure verification for prosodic naturalness. As depicted in Table 1, while MOS scores for short sentences (MOS-S) remain similar across all models, our model significantly outperforms the baseline models in OOD settings (MOS-L). These results are further validated via inter-model comparisons (Figure 3). Moreover, to assess synthesis robustness, we incorporate the widely used speech recognition WER metric. While WER is comparable for all three models in the short utterance setting, TaKOtron2-Pro excels in OOD settings as seen in its low WER scores.

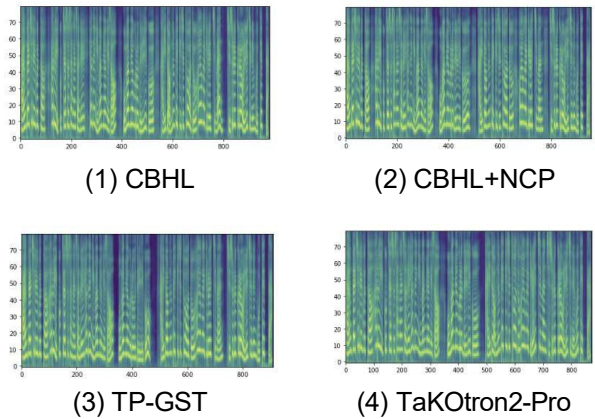

(1) CBHL     (2) CBHL+NCP

(3) TP-GST     (4) TaKOtron2-Pro

Figure 4: Mel-spectrograms with identical text input. Pauses are located in areas of dark blue.

The impact of syntactic and acoustic information on synthesized speech are independently explored and shown in Table 2. Compared to the models that explicitly incorporate syntactic features, the acoustic model tends to have higher MOS and ABX preference, as well as lower WER scores. Moreover, while the overall MOS, ABX, and WER scores are similar for the two syntactic models, there is a slightly higher preference for the CBHL+NCP model, which indicates that both local and global contextual information is useful for more natural speech synthesis. Yet, full incorporation of syntactic and acoustic attributes results in the highest scores and robustness. We further provide the visual representations of the distinct pausing patterns that arise from different component integration in Figure 4. It is worth noting that syntactic and acoustic models differ in the pause frequencies and lengths produced. For example, the CBHL+NCP syntactic model produced very frequent, but short pauses. This can be attributed to text input fragmentation into smaller prosodic units due to local and global contextual processing. In contrast, the TP-GST acoustic model produces infrequent yet longer pauses as it primarily identifies major pause breaks without utilizing local nor global syntactic information. Given that TaKOtron2-Pro capitalizes on both syntactic and acoustic cues, it exhibits assimilated pausing patterns that are observed in other single-attribute models.

## 4. Conclusion

Due to the linguistic differences and limited training resources, serious pausing errors could be noted in Korean speech when using conventional TTS models. By taking ideas from human speech production and perception, we explored how local and global contextual information, as well as acoustic characteristics could affect synthetic pause realization. Evaluations and ablation studies justified that the integration of both syntactic and acoustic information is effective in capturing pause characteristics. In fact, pauses are accurately realized even when the inferred text was much longer and complicated than the ones in the training corpus. Compared to other models, the proposed TaKOtron2-Pro demonstrates profound improvements[3].

## 5. Ethical Considerations

The task of speech synthesis, which involves generating speech for any given text, places an emphasis on achieving a high level of naturalness akin to human speech. However, this pursuit of naturalness raises important ethical considerations. One major concern is the potential for the creation of deepfakes and impersonation, where synthesized speech could be used to manipulate or deceive individuals, leading to a breach of trust. Therefore, responsible development, deployment, and usage of speech synthesis technologies are crucial to mitigate these ethical challenges.

## 6. Acknowledgements

This work was supported by the MSIT(Ministry of Science and ICT), Korea, under the ITRC(Information Technology Research Center) support program(IITP-2024-2020-0-01789) supervised by the IITP(Institute for Information & Communications Technology Planning & Evaluation), and the Technology Innovation Program(20015007, Development of Digital Therapeutics of Cognitive Behavioral Therapy for treating Panic Disorder) funded By the Ministry of Trade, Industry & Energy(MOTIE, Korea).

## 7. Bibliographical References

---

[3]Samples can be found here: https://jinny1208.github.io/demos/Project1-TaKOtron2-Pro/

## 8. Language Resource References